\documentclass[10pt,twocolumn,letterpaper]{article}

\usepackage{iccv}
\usepackage{times}
\usepackage{epsfig}
\usepackage{graphicx}
\usepackage{amsmath}
\usepackage{amssymb}
\usepackage{algorithm}
\usepackage{algpseudocode}
\usepackage{multirow}
\usepackage{pifont}% http://ctan.org/pkg/pifont
\usepackage{xcolor}
\usepackage{bbm}
\usepackage{bm}
\usepackage[accsupp]{axessibility}

\newcommand{\tick}{\checkmark}
\newcommand{\mc}[1]{\multicolumn{1}{c}{#1}}
\newcommand{\dv}{DeepLabv$3$}
\newcommand{\hrn}{HRNet}
\newcommand{\upr}{UPerNet}
\usepackage{booktabs} % cmdirule
\usepackage{subfig}

\usepackage[pagebackref=true,breaklinks=true,letterpaper=true,colorlinks,bookmarks=false]{hyperref}

\def\iccvPaperID{10145} % *** Enter the ICCV Paper ID here

\iccvfinaltrue
\ificcvfinal\fi

\newif\ifdraft
\drafttrue

\definecolor{orange}{rgb}{1,0.5,0}

\ifdraft
 \usepackage[normalem]{ulem}
 \newcommand{\RS}[1]{{\color{red}{\bf RS: #1}}}
 
 \newcommand{\PMN}[1]{{\color{orange}{\bf PMN: #1}}}

\else
 \newcommand{\sout}[1]{}
 \newcommand{\RS}[1]{{\color{red}{}}}
 
 \newcommand{\PMN}[1]{{\color{red}{}}}
 
\fi

\newcommand{\real}{\mathbb{R}}
\newcommand{\x}{\mathbf{x}}

\newcommand{\C}{\mathcal{C}}

\newcommand{\p}{\mathbf{p}} % Categorical distr. parameter vector (L)
\newcommand{\Ph}{\mathbf{\hat{P}}} % Categorical distr. parameter tensor (D,L)

\newcommand{\comment}[1]{}
\DeclareMathOperator{\EX}{\mathbb{E}}% expected value
\DeclareMathOperator*{\argmax}{arg\,max}
\algnewcommand{\LineComment}[1]{\State \(\triangleright\) #1}

\begin{document}

%%%%%%%%% TITLE
\title{Stochastic Segmentation with Conditional Categorical Diffusion Models}

\author{
Lukas Zbinden$^*$ 
\qquad 
Lars Doorenbos$^*$ 
\qquad
Theodoros Pissas\\
Adrian Thomas Huber
\qquad
Raphael Sznitman 
\qquad
Pablo Márquez-Neila
\vspace{0.5em}
\\
University of Bern, Bern, Switzerland\\
{\tt\small \{lukas.zbinden,lars.doorenbos,theodoros.pissas,raphael.sznitman,pablo.marquez\}@unibe.ch}
% \and
}

\maketitle
\ificcvfinal\thispagestyle{empty}\fi

\def\thefootnote{*}\footnotetext{Equal contribution} % \def\thefootnote{\arabic{footnote}}

%%%%%%%%% ABSTRACT
\begin{abstract}
Semantic segmentation has made significant progress in recent years thanks to deep neural networks, but the common objective of generating a single segmentation output that accurately matches the image's content may not be suitable for safety-critical domains such as medical diagnostics and autonomous driving. Instead, multiple possible correct segmentation maps may be required to reflect the true distribution of annotation maps. In this context, stochastic semantic segmentation methods must learn to predict conditional distributions of labels given the image, but this is challenging due to the typically multimodal distributions, high-dimensional output spaces, and limited annotation data. To address these challenges, we propose a conditional categorical diffusion model (CCDM) for semantic segmentation based on Denoising Diffusion Probabilistic Models. Our model is conditioned to the input image, enabling it to generate multiple segmentation label maps that account for the aleatoric uncertainty arising from divergent ground truth annotations. Our experimental results show that CCDM achieves state-of-the-art performance on LIDC, a stochastic semantic segmentation dataset, and outperforms established baselines on the classical segmentation dataset Cityscapes.

\end{abstract}

%%%%%%%%% BODY TEXT

\section{Introduction}
\label{sec:introduction}

Semantic segmentation has significantly progressed in recent years due to powerful deep neural networks. For most methods, the key objective is to generate a single segmentation output that accurately matches the image's content.
However, this may not be suitable for safety-critical domains such as medical diagnostics and autonomous driving, as images in these applications often suffer from inherent ambiguity or annotations that have differences in opinion. In these cases, generating a single coherent segmentation may be hopeless to fully describe the set of correct labeling.

\begin{figure}
    \centering
    \includegraphics[width=0.95\linewidth]{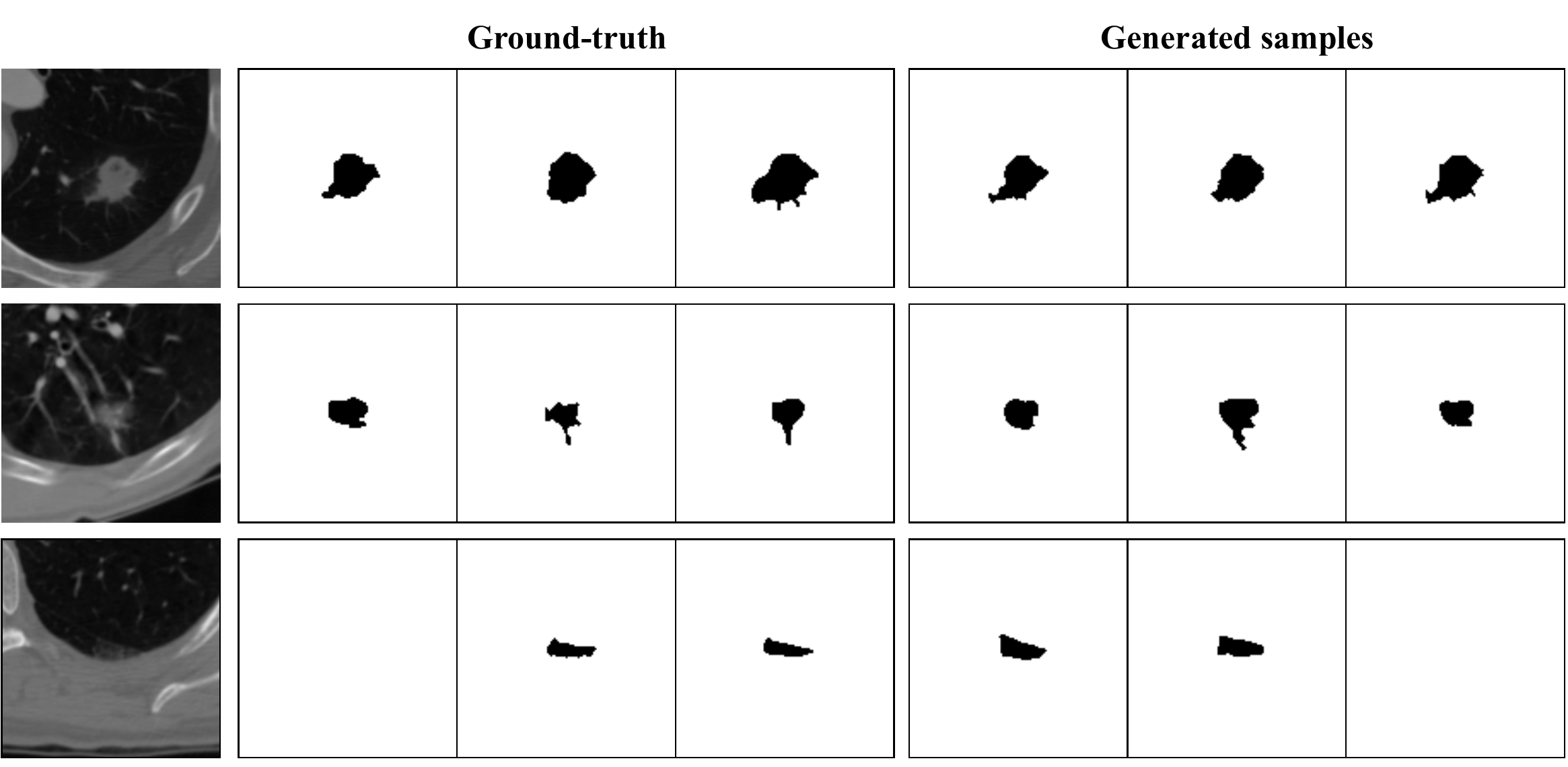}
    \caption{Examples from the LIDC dataset, where expert radiologists were asked to annotate lung nodules. Despite their expertise, they disagree significantly on many cases. Standard segmentation networks fail to capture these variations, thereby giving a false sense of confidence in model predictions. Our approach learns the distribution of possible labels, allowing us to generate realistic and diverse segmentations.}
    \label{fig:intro}
\end{figure}

Instead, multiple possible correct segmentation maps may be required to reflect the true distribution of annotations.
For instance, Fig.~\ref{fig:intro} illustrates the task of lung nodule segmentation from CT scans where expert annotators provide multiple valid segmentation maps. In this context, stochastic semantic segmentation methods must learn to predict conditional distributions of labels given the image. Doing so is challenging, however, as the distribution is typically multimodal, the output space is high-dimensional, and annotation data is limited.

Denoising Diffusion Probabilistic Models (DDPMs) appear well-suited to overcome these challenges. DDPMs have recently drawn strong interest in computer vision as a framework for learning complex distributions in high-dimensional spaces. After achieving state-of-the-art performance on image synthesis~\cite{dmsBeatGans}, they have been successfully extended to solve tasks such as text-to-image generation~\cite{saharia2022photorealistic}, counterfactual explanation generation~\cite{jeanneret2022diffusion}, inpainting ~\cite{lugmayr2022repaint}, but also image classification~\cite{zimmermann2021score} and semantic segmentation~\cite{amit2021segdiff,baranchuk2021label,wolleb2021diffusion} amongst others. 

% Problems with DDPM for segmentation
While DDPMs were originally formulated as probabilistic models able to learn high-dimensional data distributions of discrete and ordered variables (\eg,~RGB pixel values), re-formulations and modifications that allow for categorical variables (\eg,~labels)~\cite{hoogeboom2021argmax} are one of the key reasons why DDPMs are being explored in a broad range of computer vision tasks~\cite{croitoru2022diffusion}. Specifically, the ability to model the spatial distribution of categorical variables is well suited for numerous computer vision tasks, including semantic segmentation~\cite{chen2017deeplab, chen2018encoder, chu2021twins, fu2019dual, gu2022multi, harley2017segmentation, kirillov2019panoptic, li2022deep, long2015fully, zhang2022semantic, zhao2017pyramid}. Yet until now, segmentation methods using DDPMs have relied on the original discrete and ordered formulation and different heuristics to yield categorical outputs~\cite{amit2021segdiff, baranchuk2021label, wolleb2021diffusion}. Consequently, the potential advantages of adopting diffusion models of categorical variables for stochastic image segmentation are still unknown.

In light of the above, we propose a \emph{conditional categorical diffusion model}~(CCDM)~for semantic segmentation based on DDPMs, which models both the observed and the latent variables as categorical distributions. This enables the model to explicitly generate labels maps of discrete, unordered variables, thereby circumventing the need for switching between continuous and discrete domains, as in previous methods. The model is conditioned to the input image, making it possible to generate multiple segmentation label maps that account for the aleatoric uncertainty arising from image ambiguity. 
We show experimentally that our approach achieves state-of-the-art performance on LIDC, a stochastic semantic segmentation dataset, according to several performance measures. Moreover, when applied to the classical segmentation dataset Cityscapes, our method provides competitive results, outperforming established baselines.

In summary, our main contributions are the following: 
\begin{itemize}
    \item We propose a conditional categorical diffusion model capable of learning the label distribution given an input image that can be used to produce diverse segmentation samples that capture aleatoric uncertainty.
    \item For the task of learning a multi-rater semantic segmentation label distribution, our method achieves state-of-the-art performance on LIDC, being the first diffusion-based approach proposed for this task.
    \item We report competitive performance on a challenging semantic segmentation task, Cityscapes, outperforming several established baselines using a lightweight model that also leverages an off-the-shelf pre-trained feature extractor. 
\end{itemize}
%-------------------------------%
%-------------------------------%

\section{Related work}

{\bf Stochastic segmentation:} Methods for stochastic semantic segmentation aim at capturing the aleatoric uncertainty and inherent unpredictability of the labels used for segmentation. Different frameworks have been proposed to yield segmentations according to the underlying label distribution. 

Initial works aimed at equipping a standard U-Net~\cite{ronneberger2015u} with a probabilistic element to generate multiple predictions for the same image, typically accomplished by adding a conditional variational autoencoder (cVAE)~\cite{sohn2015learning}, where the low-dimensional latent space of the cVAE encodes the possible segmentation variants. In \cite{Kohl2018-hp}, samples from this latent space are upscaled and concatenated at the last layer of the U-Net. Multiple methods extend this set-up to a hierarchical version ~\cite{baumgartner2019phiseg,kohl2019hierarchical,zhang2022a-probabilistic}. Other works use normalizing flows to allow for a more expressive distribution than the Gaussian distribution in the cVAE~\cite{selvan2020uncertainty,valiuddin2021improving}, switch to a discrete latent space~\cite{qiu2020modal}, or add variational dropout and use the inter-grader variability directly as a training target~\cite{hu2019supervised}.

Several other methods do not rely on the probabilistic U-Net. Monteiro \emph{et al.}~\cite{monteiro2020stochastic} propose a network that uses a low-rank multivariate normal distribution to model the logit distribution. Kassapis \emph{et al.}~\cite{kassapis2021calibrated} leverage adversarial training to learn possible label maps based on the logits of a trained segmentation network. Zhang \emph{et al.}~\cite{zhang2022b-pixelseg} employ an autoregressive PixelCNN to model the conditional distribution between pixels. Finally, Gao \emph{et al.}~\cite{Gao2022-zt} use a mixture of stochastic experts, where each expert network estimates a mode of the uncertainty, and a gating network predicts the probabilities that an input image is segmented by one of the experts. Our method is the first to explore the use of categorical diffusion models for stochastic segmentation.

{\bf Diffusion models:} 
Generative diffusion models~\cite{sohl2015deep} have drawn much attention following their popularization by~\cite{ho2020denoising}. Since then, diffusion models have been successfully applied to various domains, such as image generation, restoration, and super-resolution \cite{croitoru2022diffusion}. 

More central to the work presented here, a few methods have attempted to apply diffusion models to semantic segmentation. Baranchuk \emph{et al.}~\cite{baranchuk2021label} first train diffusion models to generate images, then use multilayer perceptrons (MLP) on its features to predict the class label. Other works focus on binary segmentation with conditional diffusion models~\cite{amit2021segdiff,wolleb2021diffusion}. These methods generate single-channel continuous samples conditioned on the input image and obtain binary segmentation masks by thresholding the result. Directly applying continuous diffusion is also done in ~\cite{wu2022medsegdiff,wu2023medsegdiff}. Chen \emph{et al.}~\cite{chen2022analog} generate discrete data with continuous diffusion models by encoding categorical data into bits and modeling these bits as real numbers. 

Hoogeboom \emph{et al.}~\cite{hoogeboom2021argmax} propose multinomial diffusion, a variation of diffusion models designed for categorical data. Subsequently, multinomial diffusion has been applied to discrete use cases, such as for tabular data~\cite{kotelnikov2022tabddpm}, the latent space of vector-quantized variational auto-encoders~\cite{cohen2022diffusion,hu2022global} or text~\cite{hoogeboom2021argmax}. They can also generate segmentation maps in the unconditional setting at a very small resolution ($32\times64$)~\cite{hoogeboom2021argmax}. Instead, we focus on the unexplored conditional case and demonstrate results at significantly higher resolutions (up to $256\times512$).

\section{Method}

We now introduce our approach by first framing the problem setting and defining the necessary notation. We then describe categorical diffusion models and the conditioning procedure to produce stochastic semantic segmentation via diffusion. 

\subsection{Background and notation}

A denoising diffusion probabilistic model~(DDPM) is a latent variable model~$p_\theta(\x_0)=\int p_{\theta}(\x_{0:T})d\x_{1:T}$ describing the distribution of an observable variable~$\x_0\in\real^D$ using a collection of~$T$ latent variables~$\{\x_t\}_{t=1}^T$ with the same dimensionality as~$\x_0$. The joint distribution is modeled as a Markov chain~$p_{\theta}(\x_{0:T})=p(\x_T)\prod_{t=1}^T p_{\theta}(\x_{t-1}\mid\x_t)$, which is commonly known as the \emph{reverse process}. The initial~$p(\x_T)$ is set to a known, tractable distribution such as the Gaussian distribution, while the transition distribution $p_{\theta}$, parameterized by~$\theta$, is the trainable component of the model. Training a DDPM aims to approximate~$p_\theta(\x_0)$ to an empirical distribution~$q(\x_0)$ defined by a collection of samples (\eg, images from the real world). To that end, training minimizes the cross-entropy between both distributions,
\begin{equation}
    \label{eq:loss}
    \min_\theta \mathbb{E}_{\x_0 \sim q(\x_0)} \left[-\log p_{\theta}(\x_0)\right],
\end{equation}
which is intractable as it requires marginalizing over the latent variables. Instead, a tractable distribution~$q(\x_{1:T}\mid \x_0)$ is introduced and used as an approximation to the intractable true posterior~$p(\x_{1:T}\mid \x_0)$ to define the evidence lower bound (ELBO),
\begin{equation}
    \label{eq:elbo}
    \log p_{\theta}(\x_0) \ge \mathbb{E}_{\x_{1:T}\sim q(\x_{1:T}\mid \x_0)} \left[ \log \dfrac{p_\theta(\x_{0:T})}{q(\x_{1:T}\mid \x_0)} \right],
\end{equation}
where the expectation is approximated by Monte Carlo sampling. The lower bound is tight when the approximate posterior~$q$ equals the real posterior. Maximizing the ELBO over samples from~$q(\x_0)$ minimizes the cross-entropy loss of Eq.~\eqref{eq:loss}.

The key difference between DDPMs and other latent variable models is that the approximate posterior~$q(\x_{1:T}\mid \x_0)$ is fixed and not learnable. DDPMs model this distribution as a Markov chain~$q(\x_{1:T}\mid \x_0)=\prod_{t=1}^T q(\x_t\mid \x_{t-1})$, known as the \emph{forward process}. The transition distribution~$q(\x_t\mid \x_{t-1})$ is chosen to be a tractable distribution that allows efficient sampling from~$q(\x_t\mid \x_0)$ for any~$t$. The only constraint in the design of a DDPM is that $q(\x_T\mid\x_0)\approx p(\x_T)$.

The original DDPM~\cite{ho2020denoising} modeled the transition distributions of the forward and the reverse processes as Gaussian with diagonal covariance matrices, and $p(\x_T)$~as a standard multivariate normal. However, these assumptions are inadequate when the elements of~$\x_0$ belong to discrete, unordered sets, as in the task of image segmentation.

\subsection{Categorical diffusion model}

We now consider the denoising diffusion formulation to learn complex distributions of discrete image labelings. The observable variable~$\x_0\in \mathcal{L}^D$ is categorical, where $D$~is the number of pixels of the image and~$\mathcal{L}=\{1,\ldots,L\}$ is the set of discrete labels that can be assigned to each pixel. Following~\cite{hoogeboom2021argmax}, we consider that all latent variables in~$\x_{1:T}$ are also categorical and that the transition distributions for the forward and reverse processes are modeled as categorical distributions. For the forward process, the transition distribution acts element-wise over the previous state~$\x_{t-1}$ to produce the parameters of the distribution for~$\x_t$ as,
\begin{equation}
    q(\x_t\mid\x_{t-1}) = \prod_{d=1}^D q(\x_t[d] \mid \x_{t-1}[d]),
\end{equation}
where $\x_t[d]$~indicates the label at time~$t$ and pixel~$d$. In the following discussion, we will use~$x_t\in\mathcal{L}$ to refer to the label of a single pixel~$d$, and we will drop the index~$d$ for clarity. The pixel-wise transition distribution~$q(x_t \mid x_{t-1})$ gives the element-wise probability of the next label given the previous label as,
\begin{equation}
    \label{eq:q_xt_given_xt-1}
    q(x_t\mid x_{t-1}) = \C\left(x_t ; \frac{\beta_t}{L}\mathbf{1} +
    (1 - \beta_t) \mathbf{e}_{x_{t-1}}\right),
\end{equation}
where $\mathbf{1}=(1, \ldots, 1)^T$, $\mathbf{e}_\ell$ is the one-hot encoding vector with 1~in position~$\ell$ and $0$~elsewhere, and the hyperparameter~$\alpha_t = 1-\beta_t\in (0, 1)$ indicates the probability of keeping the label unchanged. $\C(x; \p)$~denotes the categorical distribution with parameter vector~$\p\in[0,1]^L$. From the properties of categorical distributions, $\C(x\mid \p) = \p[x]$ and $\sum_x \p[x] = 1$.

The transition distribution of the forward process can be composed as,
\begin{equation}
    \label{eq:q_xt_given_x0}
    q(x_t\mid x_0) = \C\left(x_t ; \dfrac{1 - \bar{\alpha}_t}{L}\mathbf{1} + \bar{\alpha}_t \mathbf{e}_{x_0}\right)
\end{equation}
with $\bar{\alpha}_t = \prod_{\tau=1}^{t} \alpha_\tau$, which enables efficient sampling of elements from the Markov chain at any location~$t$. Finally, the posterior of the transition distribution
can be computed with the previous formulas by applying Bayes rule,
\begin{equation}
    q(x_{t-1}\mid x_t, x_0) = \C\left(x_{t-1} ; \bm{\pi}(x_t, x_0)\right),
\end{equation}
with,
\begin{equation}
    \label{eq:q_params}
    \bm{\pi}(x_t, x_0) = \dfrac{1}{\tilde{\pi}}
    \left(\frac{\beta_t}{L}\mathbf{1} + \alpha_t \mathbf{e}_{x_t}\right) \odot 
    \left(\frac{1 - \bar{\alpha}_{t-1}}{L}\mathbf{1} + \bar{\alpha}_{t-1} \mathbf{e}_{x_0}\right)
\end{equation}
and $\tilde{\pi} = \frac{1 - \bar{\alpha}_t}{L} + \bar{\alpha}_t\cdot \delta_{x_t}^{x_0}$, where $\delta$~is the Kronecker delta.

The transition distribution of the reverse process is also an element-wise categorical distribution,
\begin{equation}
    \label{eq:reverse_transition}
    p_\theta(\x_{t-1}\mid\x_t) = \prod_{d=1}^D \C(x_{t-1} ; \hat{\p}_{t-1}),
\end{equation}
% old version
% where $x_{t-1}=\x_{t-1}[d]$ is the label at pixel~$d$ and $\hat{\x}_{t-1} = \hat{\X}_{t-1}(\x_t)[d]$ is the estimated parameter vector for that pixel. Unlike the forward process, the parameter vector for the pixel~$d$ is not computed considering the element~$d$ of~$\x_t$ only. Instead, the function~$\hat{\X}_{t-1}:\mathcal{L}^D \to [0, 1]^{D\times L}$ considers the entire label map~$\x_t$ to produce a collection of $D$~probability distributions, thus incorporating context in this process.
% v2
% where $x_{t-1}=\x_{t-1}[d]$ and $\hat{\p}_{t-1} = \hat{\X}_{t-1}(\x_t)[d]$ are the label and the estimated parameter vector, respectively, at pixel~$d$. Unlike the forward process, the parameter vector for the pixel~$d$ is not computed considering only the element~$d$ of~$\x_t$. Instead, the function~$\hat{\X}_{t-1}:\mathcal{L}^D \to [0, 1]^{D\times L}$, incorporates context by considering the entire label map~$\x_t$ to produce a collection of $D$~probability distributions, which we refer to as $\mathbf{\hat{P}}_{t-1} \in [0, 1]^{D\times L}$. Thus, the parameter vector for each pixel~$d$ is $\hat{\p}_{t-1} = \mathbf{\hat{P}}_{t-1}[d]$ with $\mathbf{\hat{P}}_{t-1}=\hat{\X}_{t-1}(\x_t)[d]$.
where $x_{t-1}=\x_{t-1}[d]$ and $\hat{\p}_{t-1}$ are the label and the estimated parameter vector, respectively, at pixel~$d$. Unlike the forward process, the parameter vector for the pixel~$d$ is not computed considering only the element~$d$ of~$\x_t$. Instead, it is modeled as a function $f:\mathcal{L}^D \to [0, 1]^{D\times L}$ that incorporates context by considering the entire label map~$\x_t$ to produce a collection of $D$~probability distributions for $\x_{t-1}$, which we refer to as $\mathbf{\hat{P}}_{t-1} \in [0, 1]^{D\times L}$ with $\hat{\p}_{t-1} = \mathbf{\hat{P}}_{t-1}[d]$.
% $\mathcal{L}^D \to [0, 1]^{D\times L}$
\begin{figure}
\centering
\includegraphics[width=0.47\textwidth]{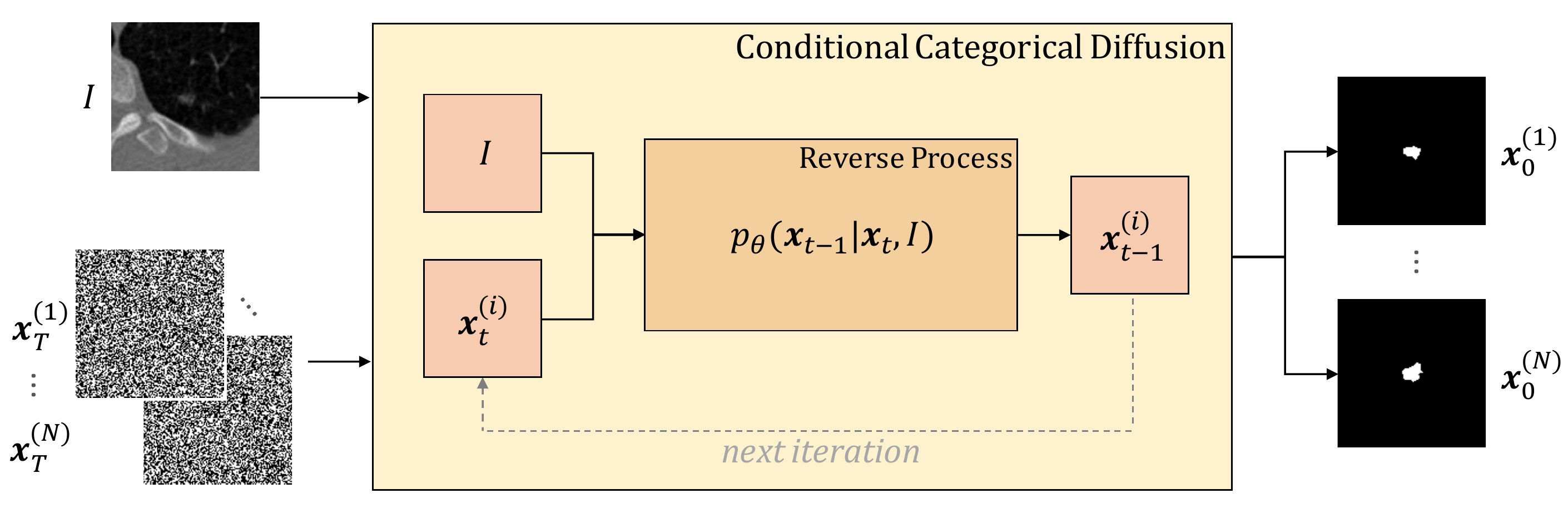}
\caption{Illustration of the reverse process of our method. The conditional categorical diffusion model (CCDM) receives as input an image~$I$ and a categorical label map $\x^{(i)}_T$ sampled from the categorical uniform noise. The reverse process of the CCDM generates a label map $\x^{(i)}_0$, which is a sample from the learned distribution~$p(\x_0\mid I)$. When repeated for $N$~samples, we obtain an empirical approximation to the multimodal label distribution for the image~$I$, learned from the annotations of multiple expert raters. }
\label{fig:ccdm_reverse_process}
\end{figure}

While it is possible to use a neural network to estimate $\Ph_{t-1}$, %to model~$\hat{\X}_{t-1}$ as neural network
Ho \etal~\cite{ho2020denoising} suggested that a consistent output space for the network led to enhanced performance. Following this idea, we train a network~$f_{\theta}$, parameterized by $\theta$, to compute $\Ph_{0} = f_{\theta}(\x_t, t) \in [0, 1]^{D\times L}$ by receiving a label map~$\x_t$ and the step~$t$. We then transform the parameter vector for each pixel, ~$\hat{\p}_0=\Ph_{0}[d]$ to the parameter vector $\hat{\p}_{t-1}$ for the same pixel of~$\x_{t-1}$ as,
\begin{align}
    &\C(x_{t-1} ; \hat{\p}_{t-1}) = \\
    &= \sum_{x_0} q(x_{t-1}\mid x_t, x_0)\cdot \C(x_0; \hat{\p}_0) \\
    &= \sum_{x_0} \C(x_{t-1}; \bm{\pi}(x_t, x_0))\cdot \C(x_0; \hat{\p}_0),
\end{align}
from which,
\begin{equation}
    \label{eq:p_params}
    \hat{\p}_{t-1} = \sum_{x_0\in\mathcal{L}} \bm{\pi}(x_t, x_0)\cdot \hat{\p}_0[x_0],
\end{equation}
where we have omitted the pixel indices~$d$ for clarity. This transformation is not necessary when~$t=1$, as then~$\hat{\p}_{t-1}=\hat{\p}_0$ computed by ~$f_{\theta}$. It is also possible to perform this computation in parallel for every pixel to efficiently obtain $\Ph_{t-1}$. Note that the result of Eq.~\eqref{eq:p_params} differs from the parameter vector computed in~\cite{hoogeboom2021argmax}, where the ill-defined expression~$\hat{\p}_{t-1}=\bm{\pi}(x_t, \hat{\x}_0)$ is employed.

% It is possible to perform this transformation for all pixels in a vectorized manner as: 
% \begin{equation}
%     \label{eq:p_params_vectorized}
%     \Ph_{t-1} = \sum_{x_0\in\mathcal{L}} \bm{\pi}(x_t, x_0)\cdot \Ph_0[:,x_0],
% \end{equation}
% \TP{need to figure out how to express this if not just skip and say we use (12) over all pixels}

\subsection{Conditional categorical diffusion}

In stochastic segmentation, the label map~$\x_0$ for an image~$I$ is modeled by a distribution~$q(\x_0\mid I)$. This distribution is often too complex to be properly approximated as a product of pixel-wise categorical distributions. We use a conditional categorical diffusion model~$p(\x_0\mid I)$ (CCDM) to model the potentially complex interactions between labels and pixels.

When conditioning the categorical diffusion model on an image, the forward process remains unchanged, $q(\x_{1:T}\mid \x_0, I) = q(\x_{1:T}\mid \x_0)$, as any latent variable is conditionally independent of the image given any previous variable. On the other hand, the reverse process needs to incorporate the dependency on the image in its transition distribution, $p_{\theta}(\x_{0:T}\mid I)=p(\x_T\mid I)\prod_{t=1}^T p_{\theta}(\x_{t-1}\mid\x_t, I)$. In practice, this dependency is enforced by an additional input to the neural network~$f_{\theta}(\x_t, t, I)$.%$\hat{\X}_0(\x_t, t, I)$.

\subsection{Training}

Training is performed by maximizing the ELBO of Eq.~\eqref{eq:elbo}. Reorganizing terms and distributing expectations for variance reduction, we express the ELBO as a sum of three terms:
\begin{align}
    & \log p_\theta(\x_0\mid I) \ge \nonumber \\
    & \mathbb{E}_{\x_1\sim q(\x_1\mid \x_0)}[\log p_\theta(\x_0\mid \x_1, I)] \\
    & - \sum_{t=2}^T \mathbb{E}_{\x_t\sim q(\x_t | \x_0)}[KL(q(\x_{t-1}| \x_t, \x_0) \| p_\theta(\x_{t-1}| \x_t, I))] \\ 
    & - KL(q(\x_T\mid \x_0) \| p(\x_T\mid I)). \label{eq:elbo_third}
\end{align}

The first two terms can be optimized by standard gradient ascent. We approximate the expectations with Monte Carlo sampling with a single sample. The sum over the time variable~$t$ is also approximated by a single uniform sample over~$\{1,\ldots,T\}$. The KL divergence of the second term is the sum of pixel-wise KL~divergences,
\begin{equation}
    \label{eq:kl_div}
    KL(q\| p) = \sum_{d=1}^D KL(q(x_{t-1}|x_t, x_0) \| p_\theta(x_{t-1}|\x_t, I)),
\end{equation}
where the parameter vectors of distributions~$q$ and~$p$ are computed with Eqs.~\eqref{eq:q_params} and~\eqref{eq:p_params}, respectively. 
Alg.~\ref{alg:training} shows the complete training procedure.

The third term of Eq.~\eqref{eq:elbo_third} does not depend on the learnable parameters~$\theta$ and is ignored during training. It is optimized by the design of the categorical diffusion model. Since the forward process converges as
\begin{equation}
    \lim_{t\to\infty} q(x_t\mid x_0) = \C\left(x; \frac{\mathbf{1}}{L}\right),
\end{equation}
we fix~$p(\x_T\mid I)$ to the element-wise uniform distribution,
\begin{equation}
    p(x_T\mid I) = p(x_T) = \C\left(x_T; \frac{\mathbf{1}}{L}\right).
\end{equation}
This ensures that~$p(\x_T\mid I)\approx q(\x_T\mid \x_0)$, making the third term of the ELBO close to zero.

% submitted
% \begin{algorithm}[t!]
% \caption{Training a CCDM with $T$~steps}
% \label{alg:training}
% \begin{algorithmic}
% \Require Training data expressed as the empirical distribution $q(\x_0, I) = q(\x_0\mid I)q(I)$.
% \Repeat
% \State $t \sim \textrm{Uniform}(\{1,...,T\})$
% \State $I \sim q(I)$
% \State $\x_0$ $\sim$ $q(\x_{0}\mid I)$\;
% \State $\x_t \sim q(\x_{t}|\x_{0})$
% \Comment{Eq.~\eqref{eq:q_xt_given_x0}}
% \State $\hat{\X}_0 \gets \hat{\X}_0(\x_t, I, t)$
% \If{$t > 1$}
%     \LineComment{Compute KL with Eqs.~\eqref{eq:kl_div}, \eqref{eq:q_params}, and~\eqref{eq:p_params}:}
%     \State $\ell \gets KL(q(\x_{t-1}|\x_{t},\x_{0})\|p_\theta(\x_{t-1}|\x_t,I))$ 

% \Else
%     \State $\ell \gets -\log p_\theta(\x_0\mid \x_1, I) = -\sum_d \log \C(x_0\mid \hat{\x}_0)$
% \EndIf
% \State $\theta \gets \theta - \nabla_\theta \ell$ \Comment{Gradient descent}
% \Until{converged}
% \end{algorithmic}
% \end{algorithm}

% new
\begin{algorithm}[t!]
\caption{Training a CCDM with $T$~steps}
\label{alg:training}
\begin{algorithmic}
\Require Training data expressed as the empirical distribution $q(\x_0, I) = q(\x_0\mid I)q(I)$.
\Repeat
\State $t \sim \textrm{Uniform}(\{1,...,T\})$
\State $I \sim q(I)$
\State $\x_0$ $\sim$ $q(\x_{0}\mid I)$\;
\State $\x_t \sim q(\x_{t}|\x_{0})$
% \Comment{Shape $D$} 
\State $\Ph_0 \gets f_{\theta}(\x_t, I, t)$ 
\Comment{shape $D\times L$} 
\If{$t > 1$}
    \LineComment{Pixel-wise application of Eq.~\eqref{eq:p_params}}
    \State $\hat{\p}_{t-1} \gets \sum_{x_0\in\mathcal{L}} \bm{\pi}(x_t, x_0)\cdot \hat{\p}_0[x_0]$  
    \Comment{shape $L$}
    % \State $\Ph_{t-1}$
    % \Comment{Eq.~\eqref{eq:p_params}} 
    \LineComment{Compute KL with Eq.~\eqref{eq:reverse_transition} and \eqref{eq:kl_div}}
    
    \State $\ell \gets KL(q(\x_{t-1}|\x_{t},\x_{0})\|p_\theta(\x_{t-1}|\x_t,I))$ 

\Else
    % \LineComment{$\forall $ pixel $d$}
    % \State $\ell_d \gets \C(x_0\mid \hat{\x}_0[d]) \quad \forall d$
    % \State $\ell \gets -\log p_\theta(\x_0\mid \x_1, I) = -\sum_d \log \C(x_0\mid \hat{\x}_0[d])$
    \State $\ell \gets -\sum_d \log \C(x_0\mid \Ph_0[d])$

\EndIf
\State $\theta \gets \theta - \nabla_\theta \ell$ \Comment{Gradient descent}
\Until{converged}
\end{algorithmic}
\end{algorithm}

At inference, the CCDM samples from~$p(\x_0\mid I)$ to generate label maps for a given image~$I$, which is achieved by traversing the Markov chain of the reverse process as outlined in Alg.~\ref{alg:segmentation} and illustrated in Fig.~\ref{fig:ccdm_reverse_process}. To minimize the noise of the generated label maps, the CCDM selects the label with maximum probability instead of sampling from~$\C(x_0\mid \hat{\p}_0)$ in the final step.

% submitted
% \begin{algorithm}[ht!]
% \caption{Inference from a CCDM with $T$~steps}
% \label{alg:segmentation}
% \begin{algorithmic}[0]
% \Require Input image~$I$
% \State $\x_{T}$ $\sim$ $\C^D\left(x_T; \frac{\mathbf{1}}{L}\right)$\; 
% \State $\x_\textrm{prev}$ $\leftarrow$ $\x_{T}$
% \For{$t = T,...,1$}
%     \State $\hat{\X}_0$ $\leftarrow$ $\hat{\X}_0(\x_\textrm{prev}, I, t)$\;
%     \If{$t > 1$}
%         \State $\hat{\x}_{t-1} \gets \sum_{x_0} \bm{\pi}(x_t, x_0)\cdot \hat{\x}_0[x_0] \quad \forall d$   \Comment{Eq.~\eqref{eq:p_params}}
%         \State $\x_\textrm{prev} \sim \prod_d \C(x_{t-1}\mid \hat{\x}_{t-1})$
%     \Else
%         \State $\x_0 \gets \argmax_{\x_0} \hat{\X}_0[:,\x_0]$
%     \EndIf
% \EndFor{}
% \end{algorithmic}
% \end{algorithm}

% new
\begin{algorithm}[ht!]
\caption{Inference from a CCDM with $T$~steps}
\label{alg:segmentation}
\begin{algorithmic}[0]
\Require Input image~$I$, $f_{\theta}$ a network trained with Alg.~\ref{alg:training}
\State $\x_{T}$ $\sim$ $\C^D\left(x_T; \frac{\mathbf{1}}{L}\right)$\; 
\State $\x_\textrm{prev}$ $\leftarrow$ $\x_{T}$ \Comment{Stores interm. and final prediction}
\For{$t = T,...,1$}
    \State $\Ph_0$ $\leftarrow$ $f_{\theta}(\x_\textrm{prev}, I, t)$\;
    \If{$t > 1$}
        \LineComment{Pixel-wise application of Eq.~\eqref{eq:p_params}}
        \State $\hat{\p}_{t-1} \gets \sum_{x_0\in\mathcal{L}} \bm{\pi}(x_t, x_0)\cdot \hat{\p}_0[x_0]$  
        \State $\x_\textrm{prev} \sim \prod_d \C(x_{t-1}\mid \hat{\p}_{t-1})$
    \Else
        \LineComment{Final prediction} %$\forall $ pixel $d$}
        \State $\x_\textrm{prev} \gets \argmax_{x_0\in\mathcal{L}} \Ph_0[:,x_0]$
        % \State $\x_\textrm{prev} \gets \mathhop{\argmax}\limits_{\x_0\in\mathcal{L}} \Ph_0[:,\x_0]$

    \EndIf
\EndFor{}
\end{algorithmic}
\end{algorithm}

\subsection{Architecture of $f_{\theta}$}%$\hat{\X}_0$}
\label{subsec:conditioning}
As described above, the neural network~$f_{\theta}$ %$\hat{\X}_0$
receives a label map~$\x_t$, a time step~$t$, and an image~$I$ to estimate the probability parameters for~$\x_0$. Its base design is a U-Net-like architecture~\cite{dmsBeatGans} with self-attention modules at the three innermost layers of the encoder and the decoder~\cite{dmsBeatGans}. The network processes the input label map represented as a binary tensor with $L$~channels encoding the label of each pixel as a one-hot vector. Parameters of the network are shared for all values of~$t$. The step variable $t$~is encoded with the standard transformer sinusoidal position embedding~\cite{ho2020denoising} and concatenated as additional channels to the input tensor and to the feature maps of intermediate layers. Similarly, information from the input image~$I$ is presented to the network as raw pixel values concatenated to the input tensor as additional channels. In some experiments we used a pre-trained transformer architecture Dino-ViT~\cite{Dino} to extract informative visual features from the image~$I$. In those cases, the extracted features were concatenated to the feature map of the third level of the U-Net encoder, which corresponds to a spatial shape equal to $\frac{1}{8}$ the shape of the input image.
\section{Experiments}

In all our experiments, we set~$T=250$ and the collection of $\beta_t$ are set following the cosine schedule proposed in \cite{nichol2021improved}. We evaluate our method on two tasks described below.

\begin{table*}[ht]
\begin{center}
\resizebox{\textwidth}{!}{
\begin{tabular}{l|cccccc|cccc}
\toprule
 & \multicolumn{6}{c}{\textbf{LIDCv1}} & \multicolumn{4}{c}{\textbf{LIDCv2}} \\
\textbf{Method} & GED$_{16}$ & GED$_{32}$ & GED$_{50}$ & GED$_{100}$ & HM-IoU$_{16}$ & HM-IoU$_{32}$ & GED$_{16}$ & GED$_{50}$ & GED$_{100}$ & HM-IoU$_{16}$\\
\hline
Prob. Unet~\cite{Kohl2018-hp} & 0.310\tiny{$\pm$0.01}$^-$ & 0.303\tiny{$\pm$0.01}$^+$ & - & 0.252\tiny{$\pm$0.004}$^\dagger$ & 0.552\tiny{$\pm$0.00}$^-$ & 0.548\tiny{$\pm$0.00}$^+$ & 0.320\tiny{$\pm$0.030}$^\ddagger$ & - & 0.252\tiny{$\pm$}$^\ddagger$ & 0.500\tiny{$\pm$0.030}$^\ddagger$\\
HProb. Unet~\cite{kohl2019hierarchical} & 0.270\tiny{$\pm$0.01}$^-$ & - & - & - & 0.530\tiny{$\pm$0.01}$^-$ & - & 0.270\tiny{$\pm$0.010}$^\ddagger$ & - & - & 0.530\tiny{$\pm$0.01}\\
PhiSeg~\cite{baumgartner2019phiseg} & 0.262\tiny{$\pm$0.00}$^-$ & 0.247\tiny{$\pm$0.00}$^+$ & - & 0.224\tiny{$\pm$0.004}$^\dagger$ & 0.586\tiny{$\pm$0.00}$^-$ & 0.595\tiny{$\pm$0.00}$^+$ & - & - & - & - \\
SSN~\cite{monteiro2020stochastic} & 0.259\tiny{$\pm$0.00}$^-$ & 0.243\tiny{$\pm$0.01}$^+$ & - & 0.225\tiny{$\pm$0.002} & 0.558\tiny{$\pm$0.00}$^-$ & 0.555\tiny{$\pm$0.01}$^+$ & - & - & - & - \\
cFlow~\cite{selvan2020uncertainty} & - & 0.225\tiny{$\pm$0.01}$^+$ & - & - & - & 0.584\tiny{$\pm$0.00}$^+$ & - & - & - & - \\
CAR~\cite{kassapis2021calibrated}  & - & - & - & 0.228\tiny{$\pm$0.009} & - & - & 0.264\tiny{$\pm$0.002} & \underline{0.248}\tiny{$\pm$0.004} & \underline{0.243}\tiny{$\pm$0.004} & \underline{0.592}\tiny{$\pm$0.005}\\
JProb. Unet~\cite{zhang2022a-probabilistic} & - & 0.206\tiny{$\pm$0.00} & - & - & - & \textbf{0.647}\tiny{$\pm$0.01} & 0.262\tiny{$\pm$0.00} & - & - & 0.585\tiny{$\pm$0.00}\\
PixelSeg~\cite{zhang2022b-pixelseg} & 0.243\tiny{$\pm$0.01} & - & - & - & 0.614\tiny{$\pm$0.00} & - & \underline{0.260}\tiny{$\pm$0.00} & - & - & 0.587\tiny{$\pm$0.01}\\
MoSE~\cite{Gao2022-zt} & 0.218\tiny{$\pm$0.003} & - & 0.195\tiny{$\pm$0.002} & \underline{0.189}\tiny{$\pm$0.002} & \textbf{0.624}\tiny{$\pm$0.004} & - & - & - & - & - \\
% \hline
AB~\cite{chen2022analog} & \underline{0.213}\tiny{$\pm$0.001} &  \underline{0.196}\tiny{$\pm$0.002} & \underline{0.193}\tiny{$\pm$0.002} & - & 0.614\tiny{$\pm$0.001} & 0.619\tiny{$\pm$0.001} & - & - & - & - \\
CIMD~\cite{cimd2023rahman} & 0.234\tiny{$\pm$0.005} & 0.218\tiny{$\pm$0.005} & 0.210\tiny{$\pm$0.005} & - & 0.587\tiny{$\pm$0.001} & 0.592\tiny{$\pm$0.002} & - & - & - & - \\
\hline
CCDM \textit{(ours)} & \textbf{0.212}\tiny{$\pm$0.002} & \textbf{0.194}\tiny{$\pm$0.001} & \textbf{0.187}\tiny{$\pm$0.002} & \textbf{0.183}\tiny{$\pm$0.002} & \underline{0.623}\tiny{$\pm$0.002} & \underline{0.631}\tiny{$\pm$0.002} & \textbf{0.239}\tiny{$\pm$0.003} & \textbf{0.216}\tiny{$\pm$0.003} & \textbf{0.210}\tiny{$\pm$0.003} & \textbf{0.598}\tiny{$\pm$0.001}\\

\bottomrule
\end{tabular}
%}  
}
\end{center}
\caption{Quantitative results on LIDCv1 and LIDCv2, with the methods ordered by year. \textbf{Bold} and \underline{underlined} indicate best and second best per column, respectively. Our results are over 3 seeds. For GED, lower is better; for HM-IoU, higher is better. No method, including ours, uses pre-trained weights. Results for CIMD~\cite{cimd2023rahman} and AB~\cite{chen2022analog} are ours. All other scores are taken from their original papers, except ($^+$) from \cite{zhang2022a-probabilistic}, ($^-$) from \cite{zhang2022a-probabilistic}, ($^\dagger$) from \cite{monteiro2020stochastic}, $\ddagger$ from \cite{kassapis2021calibrated}.}
\label{tab:lidc}
\end{table*}

\subsection{Segmentation with multiple annotations}

\paragraph{Dataset} The Lung Image Database Consortium (LIDC)~\cite{lidc-armato2011lung} binary segmentation dataset consists of 1'018~three dimensional chest CT scans of patients with lung cancer. Lung nodules of each volume are annotated by four expert raters from a pool of~$12$, yielding large differences in annotations in some cases. We extract nodule-centered slices from the CT volumes and treat each slice as an independent image.

While LIDC is the standard benchmark of stochastic segmentation methods to date (\eg~\cite{baumgartner2019phiseg,Gao2022-zt,hu2019supervised,kassapis2021calibrated,Kohl2018-hp,kohl2019hierarchical,monteiro2020stochastic,selvan2020uncertainty,zhang2022a-probabilistic,zhang2022semantic}), experimental configurations (pre-processing, training/validation/test splits, metrics) vastly differ across the literature.
We conduct our experiments on the two most prominent LIDC splits and report results on both separately. The first, referred to as LIDCv1, is used in \cite{baumgartner2019phiseg,Gao2022-zt,monteiro2020stochastic,zhang2022a-probabilistic}. LIDCv1 comprises 15'096~slices, divided into training, validation, and testing sets with the ratio $60:20:20$. The second, LIDCv2, is used in~\cite{kassapis2021calibrated,Kohl2018-hp} and contains 12'816 images with the ratio $70:15:15$.

\paragraph{Metrics} We measure the performances with the Generalised Energy Distance (GED) and the Hungarian-Matched Intersection over Union (HM-IoU)~\cite{Gao2022-zt, kassapis2021calibrated, kohl2019hierarchical}. 
Both metrics measure the difference between the distributions of generated and ground-truth label maps. We denote the metrics computed with $n$~samples using a subscript, \ie,~GED\textsubscript{$n$} and HM-IoU\textsubscript{$n$}, and we set $n$ to common values found in the literature. Note that higher number of samples yield more precise estimates.

\paragraph{Baselines} We compare our approach to eleven recent stochastic segmentation methods: probabilistic U-Net (Prob.~Unet)~\cite{Kohl2018-hp}, hierarchical probabilistic U-Net (HProb.~Unet)~\cite{kohl2019hierarchical}, PhiSeg~\cite{baumgartner2019phiseg}, stochastic segmentation network (SSN)~\cite{monteiro2020stochastic}, conditional normalizing flow (cFlow)~\cite{selvan2020uncertainty}, calibrated adversarial refinement (CAR)~\cite{kassapis2021calibrated}, joint probabilistic U-Net (JProb. Unet)~\cite{zhang2022a-probabilistic}, PixelSeg~\cite{zhang2022b-pixelseg}, mixture of stochastic experts (MoSE)~\cite{Gao2022-zt}, analog bits (AB)~\cite{chen2022analog}, and collectively intelligent medical diffusion (CIMD)~\cite{cimd2023rahman}.

Following standard practice, we use random horizontal and vertical flipping and random rotations of $0^\circ$, $90^\circ$, $180^\circ$ and~$270^\circ$ for data augmentation. The resolution of the input images is $128\times128$. We trained our method with the Adam optimizer~\cite{kingma2014adam} until convergence of the GED metric on the validation set, a polynomial learning rate scheduling starting from $1e^{-4}$ and ending with $1e^{-6}$, and batch size of~64. We applied Polyak averaging with $\alpha =  0.99995$.

\subsection{Segmentation with a single annotation}
\label{sec:single_annotation}

We also evaluate our method with Cityscapes, a classical multi-class segmentation dataset where each image is annotated with a single label map. It comprises 2'975~RGB images of urban scenes for training and 500~images for validation, with each image labeled using 19~possible classes.

We compare our approach to several established baselines using the validation set: DeepLabv3~\cite{DeepLabv3}, HRNet~\cite{HRNet}, and UPerNet~\cite{UPerNet}, with both ResNet~\cite{ResNet} and Swin~\cite{Swin} backbones. 

Besides our standard method, which performs image conditioning by concatenating the raw pixel values as channels of the input tensor, we also included in our comparison a variant CCDM-Dino which leverages pre-trained Dino-ViT features~\cite{Dino} as additional conditioning concatenated to intermediate feature maps of our model's encoder.

Experiments are conducted separately for two different image resolutions: $128\times256$ and $256\times512$. For all reported methods, we first resize the images to a fixed resolution and then apply color jittering, random flipping, and standard ImageNet intensity normalization as data augmentation. %\LZ{don't see the latter in the tables?}. 
All baselines are trained for $500$~epochs with a batch size of~$32$, with optimizers, learning rate schedules, and weight decay settings as reported in their respective publications (reported in detail in the supplementary material). 

Our method was trained for $800$~epochs with a batch size of~$32$ at~$128\times256$ and of $16$ at~$256\times512$, using the Adam optimizer~\cite{kingma2014adam} with a learning rate of $1e^{-4}$ linearly decayed to $1e^{-6}$. We applied Polyak averaging with~$\alpha=0.999$.

Performance is measured with the mean intersection-over-union~(mIoU). Unlike GED and HM-IoU, the metric mIoU is incompatible with multiple label maps per image. During inference, CCDM generates multiple label maps per image that are subsequently fused into a single label map for performance assessment. We found that fusing by averaging the predicted probabilities resulted in superior performances compared to fusing by majority vote.

\begin{figure*}
    \centering
    \includegraphics[width=0.95\linewidth]{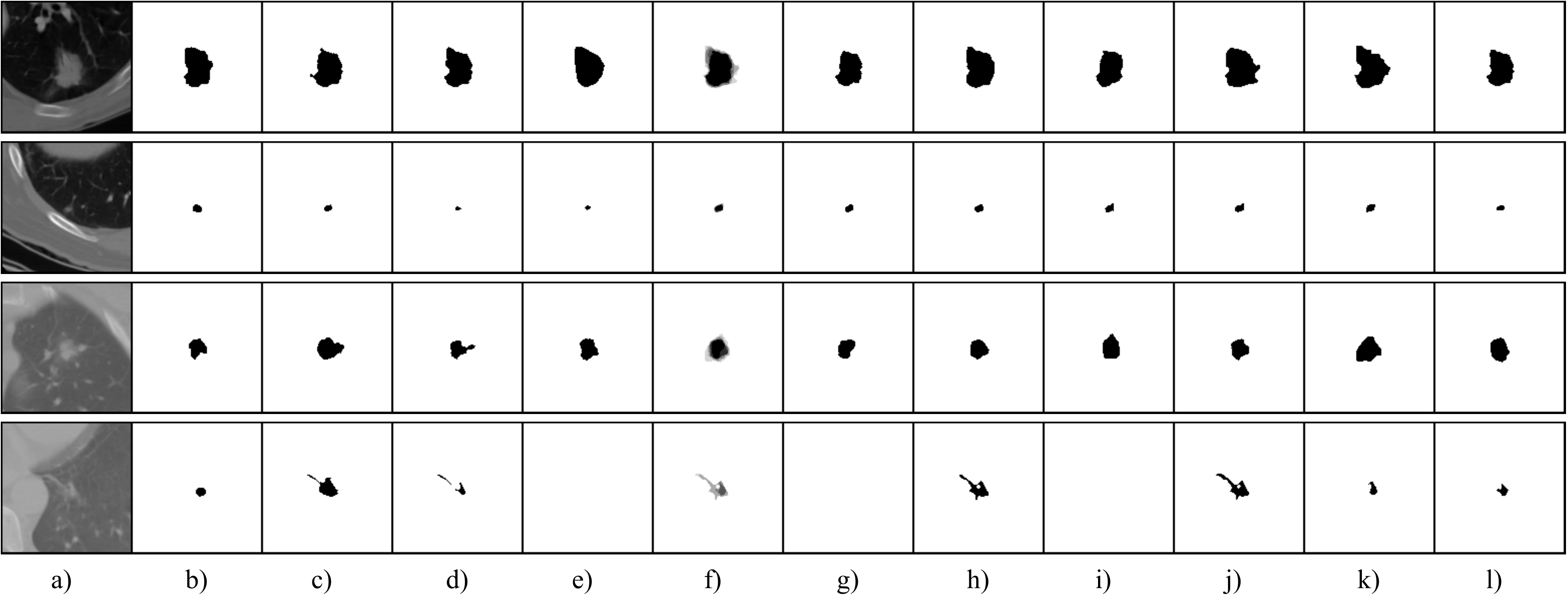}
    \caption{Qualitative results on four LIDC images with our method. (a) shows the image, (b)-(e) its four labels, (f) the mean prediction of our CCDM over six predictions, and (g)-(l) six individual predictions.}
    \label{fig:quali_lidc}
\end{figure*}

\section{Results}
\label{sec:results}

\subsection{LIDC}
We report performances on LIDCv1 and LIDCv2 in Tab.~\ref{tab:lidc} and qualitative results in Fig.~\ref{fig:quali_lidc}. Due to the lack of consistent evaluation protocols, we use a total of 10 metrics, thereby covering all the baselines and allowing for direct comparisons.

Our CCDM reaches the best performance for eight out of the ten metrics, despite its relatively small size, with 9M parameters compared to, \eg,~the 42M parameters of MoSE. CCDM also outperforms recent continuous diffusion models for segmentation, including AB~\cite{chen2022analog} (9M parameters) and CIMD~\cite{cimd2023rahman} (24M parameters). On HM-IoU\textsubscript{16}, the CCDM has a lower mean performance than MoSE by 0.001, but with only half the standard deviation. The JProb.~Unet reaches a higher HM-IoU\textsubscript{32} than all other methods, despite being considerably worse for GED\textsubscript{32} than our~CCDM. Furthermore, on LIDCv2, the JProb.~Unet achieves only the third-best score on GED\textsubscript{16}, and fourth-best on HM-IoU\textsubscript{16}. This result indicates how comparing results obtained on different LIDC versions with each other can be misleading.

Fig.~\ref{fig:quali_lidc} presents qualitative results from our method. In columns (g)-(l), we see that our CCDM generates a distribution of samples that captures the annotation variability created by the four expert raters. Further, as seen in the bottom example, the CCDM also generates empty samples according to the annotations (b)-(e).

\begin{figure}
\centering
\includegraphics[width=0.3\textwidth]{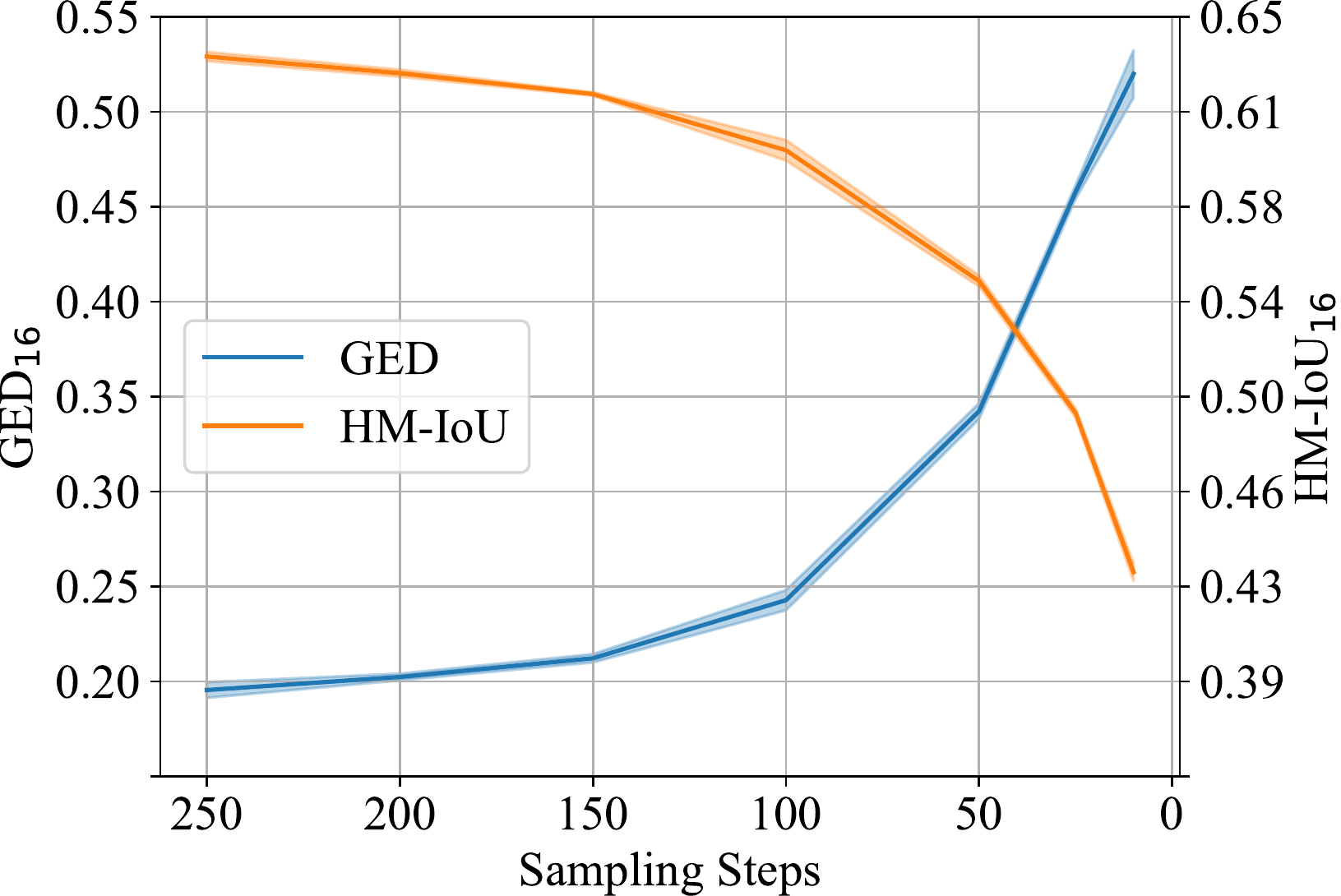}
\caption[]{LIDC GED and HM-IoU versus the number of sampling steps on LIDC. Evaluated on 500 random test images using 16 samples each, over 3 seeds.}
\label{fig:sampling_steps}
\end{figure}

\noindent \textbf{Reduced number of time steps for sampling:} During inference, traversing the $T$~steps of the reverse process makes sampling from DDPMs slow. A straightforward solution~\cite{nichol2021improved} involves traversing only a subset of nodes of the reverse process, $\{\x_{k\tau}:\tau\in\{0, \ldots, T/k\}\}$, reducing the number of steps by a factor~$k$. This technique accelerates inference at the expense of reduced performance. To illustrate the trade-offs between performance and speed, Fig.~\ref{fig:sampling_steps} presents the evolution of GED$_{16}$ and HM-IoU$_{16}$ as the number of inference steps is reduced. As expected, CCDMs perform best when the number of training and inference steps are equal, but a reasonable increase in speed without a large sacrifice in performance is possible.

\subsection{Cityscapes}

\begin{figure*}
\centering
\includegraphics[width=0.99\textwidth]{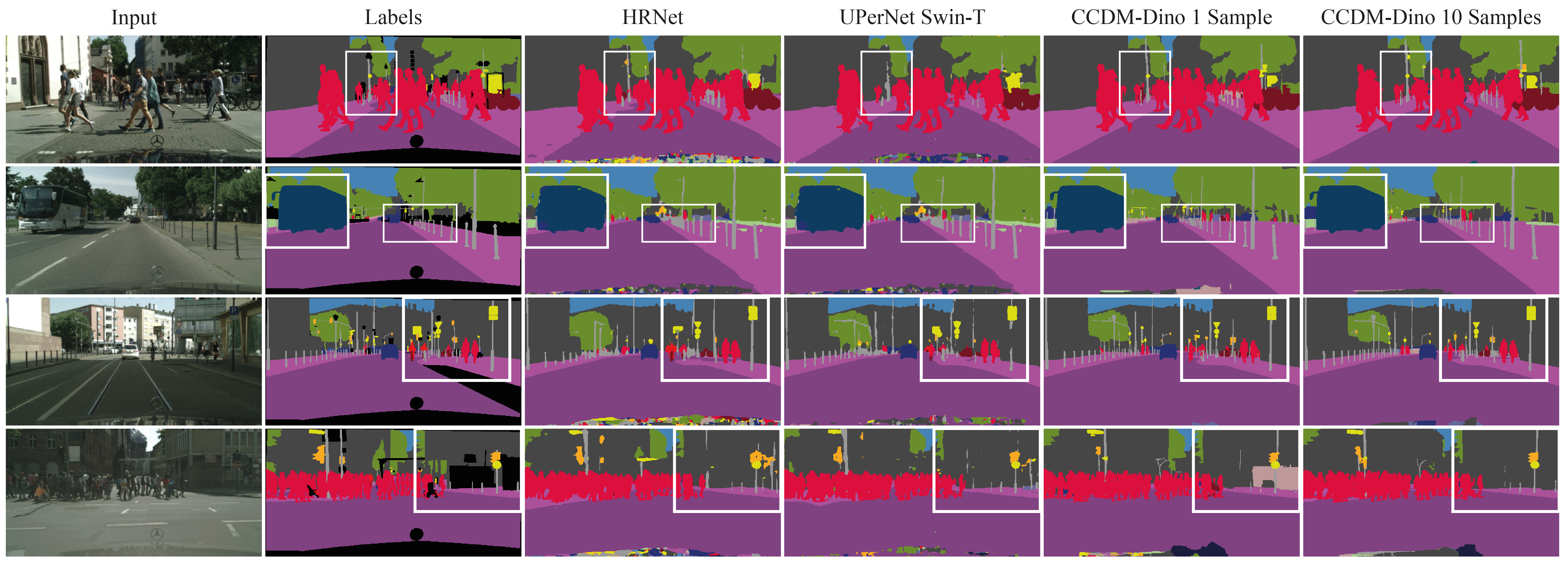}
\caption[]{Qualitative comparisons on Cityscapes. All methods are trained and tested at a resolution of $256\times512$. Our method produces structures with greater visual realism than other baselines. This is especially noticeable inside the marked regions.
}
\label{fig:quali_cts}
\end{figure*}

\begin{table}[ht]
% \centering
\resizebox{80mm}{!}{
\begin{tabular}{*5c}
    \toprule
    \multicolumn{3}{c}{\textbf{Method}}             & \multicolumn{2}{c}{mIoU final (best)} \\
    \cmidrule(r){1-3}                                 \cmidrule(l){4-5}         
    Architecture       & \mc{Backbone}   & \#params     & \mc{$128\times256$}    &  \mc{$256\times512$}\\
    \midrule
    \mc{\dv} \cite{DeepLabv3}      & \mc{ResNet50}\ (\tick)                       & 39m                  & 43.4 (44.1)     &   58.6 (59.2)       \\ 
    \mc{\dv} \cite{DeepLabv3}     & \mc{ResNet101} (\tick)                       & 58m                  & 43.8 (45.5)     &   59.2 (59.8)    \\ 
    \mc{\upr} \cite{UPerNet}     & \mc{ResNet101} (\tick)                     & 83m                 & 45.5 (47.1)      &  60.7 (61.2)     \\ 
    \mc{\hrn} \cite{HRNet}     & \mc{w48v2}  (\tick)                    & 70m                 & 48.2 (49.5)      &  63.3 (64.2)         \\ 
    \mc{\upr} \cite{Swin}      & \mc{Swin-Tiny} (\tick)                    & 58m                 & 54.2 (55.9)      &   \underline{65.5} (66.0)    \\ 

    \midrule
    \mc{CCDM} (ours) & \mc{-}               &  \quad          &  \quad        & \quad   \\  
    
    samples=1 & \mc{}               & 30m           &  53.2         & 60.3     \\  
    samples=5 & \mc{}                & 30m            &  55.4           & 62.0   \\  
    samples=10 & \mc{}               & 30m            & 56.2          &  62.4    \\  
  
    \midrule

    \mc{CCDM (ours)} & \mc{Dino ViT-S} ($\dagger$)             & \quad           &  \quad        & \quad   \\  
    samples=1 & \mc{}               & 30m + \textcolor{gray}{20M}           &  55.5         &  64.0     \\  
    samples=5 & \mc{}               & 30m + \textcolor{gray}{20M}           & \underline{56.9}           & 65.4   \\  
    samples=10 & \mc{}              & 30m + \textcolor{gray}{20M}           & \textbf{57.3}          &  \textbf{65.8}    \\  
    \bottomrule
    
\end{tabular}}

\caption{Results on Cityscapes-val for resolutions $128\times256$ and $256\times512$. \textbf{Bold} and \underline{underlined} indicate best and second best per column, respectively. (\tick) and ($\dagger$) indicate supervised and self-supervised pretraining of the backbone, respectively. \textcolor{gray}{Gray} indicates pretrained, non-finetuned parameters. We report final performance for our method and baselines. For the latter we also provide best achieved performance during training (in parenthesis). For CCDM methods, the field \emph{samples} indicates the number of generated samples for label map fusion, as explained in Sect~\ref{sec:single_annotation}.} 
\label{tab:cts}
\end{table}

\begin{table}[ht]
\centering

\resizebox{60mm}{!}{
  \begin{tabular}{*5c}
    \toprule
    \multicolumn{2}{c}{CCDM Capacity}             &       \multicolumn{3}{c}{mIoU}($128\times256$)   \\
    \cmidrule(r){1-2}                       \cmidrule(l){3-5}   
    \#params  & UNet Levels         & samples=1 & samples=5 & samples=10\\
    \midrule
    5.4M & 4          &  37.8 & 39.7 & 40.6  \\    
    7.5M & 5       &  44.7 & 48.3 & 48.5          \\
    22M  & 4       & 51.6 & 54.0 & 53.6 \\
    30M  & 5        &  \textbf{53.2} & \textbf{55.4} & \textbf{56.2}   \\    
    \bottomrule
\end{tabular}
}
% }
\caption{Effect of increasing CCDM capacity (without feature conditioning).}
\label{tab:cts_ablations}

\end{table}

Experimental comparisons on Cityscapes are presented in Tab.~\ref{tab:cts}, and qualitative examples are provided in Fig.~\ref{fig:quali_cts}. Experiments at $128\times256$ demonstrate that CCDM-Dino outperforms all other methods, even when only a single sample is used. CCDM-raw also remains competitive, being outperformed only by one baseline (UPerNet+Swin-Tiny), despite using only between $36\%$ and $51\%$~of the parameters of other models. Similarly, at $256\times512$, CCDM-Dino outperforms four of the baselines with a single sample, lags behind UPerNet+Swin-Tiny only by $0.1$~percent points with $5$~samples, and outperforms all baselines with $10$~samples. As expected, averaging across more samples improves performance for both CCDM-raw and CCDM-Dino, albeit with diminishing gains. Furthermore, the addition of Dino features boosts single-sample performance by $2.3$~percent points at $128\times256$, and $3.7$~percent points at $256\times512$, hinting the greater value of adding feature conditioning for generating segmentation at a higher resolution. 

\noindent \textbf{CCDM Capacity}: Tab.~\ref{tab:cts_ablations}(b) demonstrates the effect of increasing the capacity of CCDM. Using more U-Net feature levels, and increasing the number of parameters by doubling the number of channels per level, increases the performance regardless of the number of samples used for inference.

\section{Conclusion} 

We introduced conditional categorical diffusion models (CCDMs) that are capable of effectively modeling pixel-level semantic distributions. Notably, and contrary to standard deterministic segmentation approaches, our model can produce diverse samples given an input image, thereby capturing the aleatoric uncertainty. Our method learns a multi-modal label distribution of segmentations, induced by annotations from multiple expert raters, for which it achieves state-of-the-art results on a challenging medical imaging dataset, LIDC. Additionally, we demonstrate that it can achieve competitive performance on a standard multi-class semantic segmentation benchmark, Cityscapes, by outperforming several established, heavily engineered baselines despite using significantly fewer parameters. 

One limitation of our method is the requirement of several iterations for producing a sample, which is a common shortcoming of diffusion models. Accelerating sampling constitutes a crucial research direction, orthogonal to the present work. 
Finally, resolution scaling remains notoriously difficult for diffusion models, with successful examples relying on massive computational resources to train cascades of models that gradually increase resolution~\cite{cascaded_DDPM,Imagen} or operate on the latent space of existing embedding methods for continuous data (\eg~images) \cite{LatentDiffusionModels} that are not available for categorical data.

\section*{Acknowledgements}
This work was partially funded by the University of Bern, Swiss National Science Foundation Grants \#320030-188591, \#200021-192285, and \#200021-191983.

%------------------------------------------------------------------------
%------------------------------------------------------------------------

%%%%%%%%% REFERENCES

{\small
\bibliographystyle{ieee_fullname}
\bibliography{egbib}
}

\clearpage

\section{Supplementary material}

\subsection{Metrics details}
The GED and HM-IoU metrics used in our work are computed as follows:

{\bf GED:} Let~$p_m$ be the distribution over samples generated by a model and $p_{gt}$ the distribution over possible ground-truth labels; the GED is computed as
\begin{align}
    \label{eq:ged}
    \text{GED}(p_m, p_{gt}) = &2\EX_{s\sim p_m,\hat{s}\sim p_{gt}}[d(s,\hat{s})] - \EX_{s,\hat{s}\sim p_{gt}}[d(s,\hat{s})] \nonumber\\
    & - \EX_{s,\hat{s}\sim p_m}[d(s,\hat{s})],
\end{align}
where the distance function $d(\cdot,\cdot) = 1 - \text{IoU}(\cdot,\cdot)$. 

{\bf HM-IoU:} Finds the optimal matching between ground truth and generated samples. Specifically, for $n$ generated samples, the ground-truth samples are duplicated to $n$. Then, the HM-IoU is defined as the maximum IoU possible, given that every generated sample is matched with a unique ground-truth label, found by minimizing
\begin{equation}
    \text{HM-IoU} = \min_X\sum_i\sum_j d(i,j)X_{i,j},
\end{equation}
where $X$ is a boolean matrix that assigns every row to a unique column using $d(\cdot,\cdot) = 1 - \text{IoU}(\cdot,\cdot)$.

\subsection{Sample diversity}

Sample diversity is the expected distance between generated samples, \ie,~$\EX_{s,\hat{s}\sim p_m}[d(s,\hat{s})]$, which corresponds to the last term of GED in Eq.~\eqref{eq:ged}.
We report the sample diversity for 16, 32, 50, and 100 samples for both LIDC splits in Tab.~\ref{tab:diversity1} and Tab.~\ref{tab:diversity2}. 

\begin{table}[ht]
\begin{center}
\resizebox{0.5\textwidth}{!}{
\begin{tabular}{l|cccc}
\toprule
 & \multicolumn{4}{c}{\textbf{LIDCv1}}\\
\textbf{Method} & Div$_{16}$ & Div$_{32}$ & Div$_{50}$ & Div$_{100}$\\
\hline
CCDM & 0.491\tiny{$\pm$0.001} & 0.509\tiny{$\pm$0.001} & 0.515\tiny{$\pm$0.002} & 0.519\tiny{$\pm$0.002} \\
\bottomrule
\end{tabular}
%}  
}
\end{center}
\caption{Sample diversity for our method on LIDCv1.}
\label{tab:diversity1}
\end{table}

\begin{table}[ht]
\begin{center}
\resizebox{0.5\textwidth}{!}{
\begin{tabular}{l|cccc}
\toprule
 & \multicolumn{4}{c}{\textbf{LIDCv2}} \\
\textbf{Method} & Div$_{16}$ & Div$_{32}$ & Div$_{50}$ & Div$_{100}$ \\
\hline
CCDM & 0.487\tiny{$\pm$0.003} & 0.503\tiny{$\pm$0.003} & 0.509\tiny{$\pm$0.003} & 0.515\tiny{$\pm$0.002} \\
\bottomrule
\end{tabular}
%}  
}
\end{center}
\caption{Sample diversity for our method on LIDCv2.}
\label{tab:diversity2}
\end{table}

\subsection{Model size}
While our 9M CCDM as reported in Tab.~\ref{tab:lidc} is of comparable size to most other baselines, we show in Tab.~\ref{tab:lidc_size} that by increasing the size of our CCDM from 9M to 41M, we get an increase in performance across all six metrics computed on LIDCv1. Additionally, the CCDM seems to benefit more from the increase in size than MoSE~\cite{Gao2022-zt}. While we already outperform the other baselines with our 9M model, this result suggests that we can improve the performance even further by using larger models.
\begin{table}[ht]
\begin{center}
\resizebox{0.5\textwidth}{!}{
\begin{tabular}{lc|cccccc}
\toprule
 & \multicolumn{4}{c}{\textbf{LIDCv1}}\\
\textbf{Method} & \#params & GED$_{16}$ & GED$_{32}$ & GED$_{50}$ & GED$_{100}$ & HM-IoU$_{16}$ & HM-IoU$_{32}$\\
\hline
MoSE~\cite{Gao2022-zt} & 9m & 0.219 & - & 0.195 & 0.190 & 0.620 & - \\
MoSE~\cite{Gao2022-zt} & 42m & 0.218 & - & 0.195 & 0.189 & 0.624 & - \\
\midrule
CCDM & 9m & 0.212 & 0.194 & 0.187 & 0.183 & 0.623 & 0.631\\
CCDM & 41m & 0.207 & 0.189 & 0.182 & 0.177 & 0.629 & 0.636\\
\bottomrule
\end{tabular}
%}  
}
\end{center}
\caption{Performance of CCDM and MoSE on LIDCv1 with different model sizes.}
\label{tab:lidc_size}
\end{table}

\subsection{Training settings of baselines on Cityscapes}

On Cityscapes, all baselines were trained for $500$~epochs using the optimizer, learning rate schedule, and weight decay (denoted by~$w_d$) reported in their original publications. Tab.~\ref{tab:settings_cts} details these settings for each case. All models are trained using a cross-entropy loss.

\begin{table}[htb]
\centering
\resizebox{0.5\textwidth}{!}{
\begin{tabular}{*7c}
    \toprule
    \multicolumn{2}{c}{Method}  &  \multicolumn{5}{c}{\textbf{Settings}}                     \\
    \cmidrule(r){1-2}               \cmidrule(l){3-7}
          Arch.   & Backbone                & Lr               & Decay        & $w_{d}$          & Batch Size    & Optim \\
    \midrule
    \mc{\hrn}\cite{HRNet}     & \mc{w$48$v$2$}          & $10^{-2}$        & polynomial   & $5\times10^{-5}$ & \mc{$32$}    & sgd \\ 
    \mc{\dv}\cite{DeepLabv3}      & \mc{ResNet$50/101$}     & $10^{-2}$        & polynomial   & $5\times10^{-5}$ & \mc{$32$}    & sgd \\ 
    \mc{UPerNet}\cite{UPerNet}  & \mc{ResNet$101$}        & $10^{-2}$        & polynomial   & $5\times10^{-5}$ & \mc{$32$}    & sgd \\
    \mc{UPerNet}\cite{Swin}  & \mc{Swin-T}             & $10^{-4}$        & warmup+linear       & $10^{-2}$        & \mc{$32$}    & AdamW \\
    % \mc{UNet}     & \mc{-}                         & $10^{-4}$        & linear       & \mc{-}           & \mc{$16/32$} & Adam \\
    % \mc{UNet}     & \mc{Dino-ViT-S} ($\dagger$)    & $10^{-4}$        & linear       & \mc{-}           & \mc{$16/32$} & Adam \\
    \bottomrule
\end{tabular}
} % resizebox  
\caption{Training settings of baselines on Cityscapes.
}
\label{tab:settings_cts} 
\end{table}

\subsection{Additional comparisons on Cityscapes}

\begin{table}[ht]
% \centering
\resizebox{80mm}{!}{
\begin{tabular}{*5c}
    \toprule
    \multicolumn{3}{c}{\textbf{Method}}             & \multicolumn{2}{c}{mIoU} \\
    \cmidrule(r){1-3}                                 \cmidrule(l){4-5}         
    Architecture       & \mc{Backbone}   & \#params     & \mc{$128\times256$}    &  \mc{$256\times512$}\\
    \midrule
    \mc{UNet} (CE) \cite{dmsBeatGans} & \mc{-}                      & 30m                & 48.7       &  61.0  \\

    \midrule
    \mc{CCDM} (ours) & \mc{-}     &  \quad          &  \quad        & \quad   \\  
    
    samples=1 & \mc{}             & 30m           &  53.2         & 60.3     \\  
    samples=5 & \mc{}             & 30m            &  55.4           & 62.0   \\  
    samples=10 & \mc{}            & 30m            & 56.2          &  62.4    \\  
  
    \midrule
    \mc{UNet (CE)} \cite{dmsBeatGans} & \mc{Dino ViT-S} ($\dagger$) & 30m + \textcolor{gray}{20M}                & 53.4       &  63.2   \\ 
    \midrule

    \mc{CCDM (ours)} & \mc{Dino ViT-S} ($\dagger$)             & \quad           &  \quad        & \quad   \\  
    samples=1 & \mc{}               & 30m + \textcolor{gray}{20M}           &  55.5         &  64.0     \\  
    samples=5 & \mc{}               & 30m + \textcolor{gray}{20M}           & \underline{56.9}           &  \underline{65.4}   \\  
    samples=10 & \mc{}              & 30m + \textcolor{gray}{20M}           & \textbf{57.3}          &  \textbf{65.8}    \\  
    \bottomrule
    
\end{tabular}}

\caption{Comparison of our method to UNet and UNet-Dino, trained with standard Cross-Entropy (CE) loss, on Cityscapes-val. \textbf{Bold} and \underline{underlined} indicate best and second best per column, respectively. ($\dagger$) indicates self-supervised pretraining of the backbone. \textcolor{gray}{Gray} indicates pretrained, non-finetuned parameters.}
\label{tab:cts_unet}
\end{table}

\begin{figure*}[]
\centering
\includegraphics[width=0.99\textwidth]{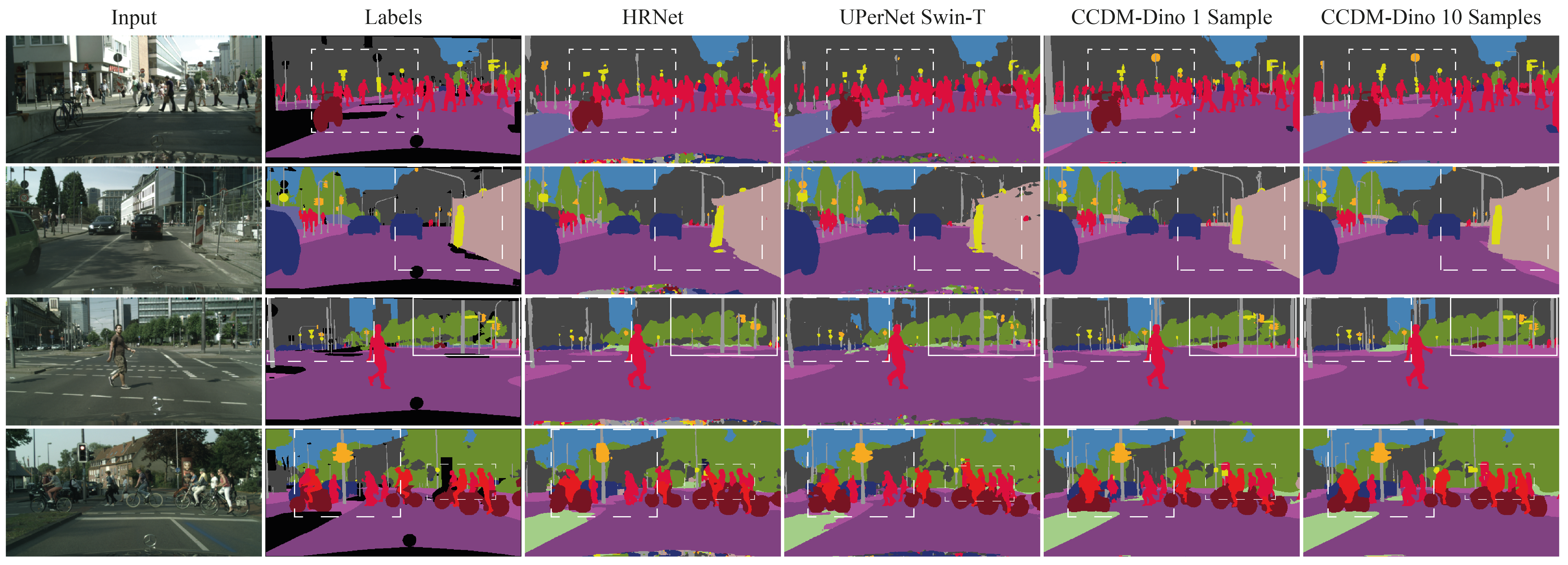}
\caption[]{Qualitative comparisons of our method to competitive baselines on Cityscapes validation set.
}
\label{fig:quali_cts_2}
\end{figure*}

\begin{figure*}[h!]
\centering
\includegraphics[width=0.99\textwidth]{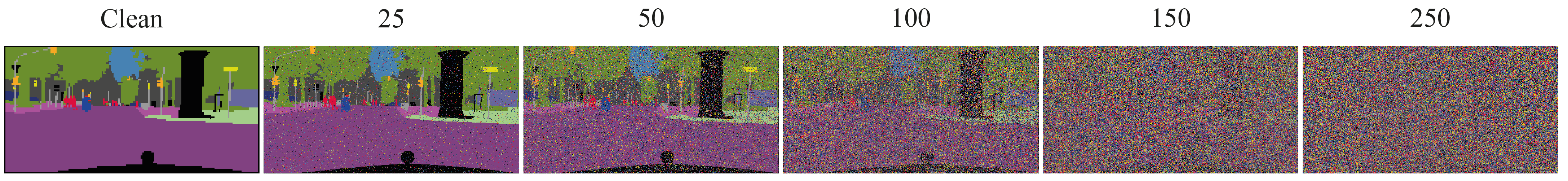}
\caption[]{Visualization of the forward diffusion process at different time steps.}
\label{fig:quali_cts_forward}
\end{figure*}

We evaluate the gains of CCDMs with respect to their backbone architectures when used as standalone segmentation models. To this end, we compare the performance of our CCDM trained as defined in Alg.~\ref{alg:training} and the UNet trained with a standard cross-entropy loss, both on the Cityscapes dataset. Similarly, we compare CCDM-Dino to its standalone backbone architecture DinoViT-S. In all cases, we adopt the same training settings as our method, namely, $800$~epochs, linearly decayed learning rate, batch size of~$32$ at $128\times256$ and $16$~at $256\times512$. As shown in Tab.~\ref{tab:cts_unet}, CCDM and CCDM-Dino outperform their respective standalone architectures.

We also provide additional qualitative comparisons of our method to competitive baselines in Fig.~\ref{fig:quali_cts_2}. Finally, Fig.~\ref{fig:quali_cts_forward} shows an example of the evolution of a Cityscapes label map under the forward diffusion process described by Eq.~\eqref{eq:q_xt_given_xt-1}.

\end{document}